\documentclass[letterpaper]{article} 
\usepackage{aaai24}  
\usepackage{fancyhdr}
\usepackage{times}  
\usepackage{helvet}  
\usepackage{courier}  
\usepackage[hyphens]{url}  
\usepackage{graphicx} 
\usepackage{natbib}  
\usepackage{caption} 
\usepackage{algorithm}
\usepackage{algorithmic}
\usepackage{multirow}
\usepackage{amssymb}
\usepackage{pdfpages}
\usepackage{newfloat}
\usepackage{listings}
\DeclareCaptionStyle{ruled}{labelfont=normalfont,labelsep=colon,strut=off} 
\floatstyle{ruled}
\newfloat{listing}{tb}{lst}{}
\floatname{listing}{Listing}
\title{Unearthing Common Inconsistency for Generalisable Deepfake Detection}
\author{
    Beilin Chu\textsuperscript{\rm 1},
    Xuan Xu\textsuperscript{\rm 1},
    Weike You\textsuperscript{\rm 1},
    Linna Zhou\textsuperscript{\rm 1}\thanks{Corresponding author.}
}
\affiliations{
    \textsuperscript{\rm 1}School of Cyberspace Security, Beijing University of Posts and Telecommunications\\
    \{beilin.chu, sh22xuxuan, ywk, zhoulinna\}@bupt.edu.cn
}

\usepackage{bibentry}

\begin{document}
\pagestyle{plain}

\maketitle

\thispagestyle{plain} 

\begin{abstract}
Deepfake has emerged for several years, yet efficient detection techniques could generalize over different manipulation methods require further research. While current image-level detection method fails to generalize to unseen domains, owing to the domain-shift phenomenon brought by CNN's strong inductive bias towards Deepfake texture, video-level one shows its potential to have both generalization across multiple domains and robustness to compression. We argue that although distinct face manipulation tools have different inherent bias, they all disrupt the consistency between frames, which is a natural characteristic shared by authentic videos. Inspired by this, we proposed a detection approach by capturing frame inconsistency that broadly exists in different forgery techniques, termed unearthing-common-inconsistency (UCI). Concretely, the UCI network based on self-supervised contrastive learning can better distinguish temporal consistency between real and fake videos from multiple domains. We introduced a temporally-preserved module method to introduce spatial noise perturbations, directing the model's attention towards temporal information. Subsequently, leveraging a multi-view cross-correlation learning module, we extensively learn the disparities in temporal representations between genuine and fake samples. Extensive experiments demonstrate the generalization ability of our method on unseen Deepfake domains.
\end{abstract}

\section{Introduction}

In recent years, the emergence of Deepfake technology has captured global attention, showcasing remarkable advancements in the field of deep learning. With its ability to manipulate and create hyper-realistic multimedia content, Deepfake techniques \cite{DeepFakes,thies2016face2face,FaceSwap,thies2019deferred} represent a significant shift in how humans interact with digital media. However, alongside its potential benefits, Deepfake also give rise to notable ethical, societal, and security concerns. For this reason, the development of reliable detection methods becomes imperative to tackle the multifaceted challenges posed by Deepfake technology.

To mitigate the threats posed by Deepfake, numerous detection methods have been proposed. Currently, these detection techniques can be broadly categorized into two types: image-level and video-level approaches. Image-level methods typically employ Deep Convolutional Neural Networks (DCNNs) as the backbone to identify subtle artifacts in pixel level \cite{dang2020detection,li2020face,liu2020global}. In specific, most of them take advantage of CNNs’ strong inductive bias towards image styles (i.e. texture), to learn pixel distribution discrepancies between authentic and synthetic images \cite{baker2018deep,geirhos2018imagenet,hermann2020origins} . As such, numerous experiments exhibit satisfying performances on several public datasets, such as FaceForensics++ \cite{rossler2019faceforensics++,li2020celeb,dolhansky2019deepfake}, Celeb-DF, and DFDC. However, related research has shown that such ability is intrinsically sensitive to unseen domains, since the style of texture may vary among manipulation methods. On the other hand, video-level approaches utilize the inconsistency between successive frames, which is caused by ignorance of inter-frame interaction in the manipulation process. Experiments from several works \cite{sabir2019recurrent} for face manipulation, Exploiting prediction error inconsistencies through lstm-based, Deepfake video detection through optical flow base, Deepfake detection using spatiotemporal convolutional network, Lips Don’t Lie: A Generalisable and Robust Approach to] have shown that such inconsistency commonly exists in different types of forgery methods, making it a potentially discriminative clue to generalize across unseen domains. However, recent video-level detectors still suffer from downgrading when tested on unseen domains. We argue that the majority of video-level detection methods solely extract temporal inconsistency from single source domain, ignoring method-invariant temporal inconsistency that broadly exists in different fake videos. 

To tackle the aforementioned issues with generalisable Deepfake detection, some 
recent works \cite{dong2023contrastive,zhao2022self} deploy self-supervised learning to address this problem. They achieved surprising performances on cross-domain tests, which promises a realistic direction of generalisable detection. Moreover, these self-supervised methods are mainly devised at image-level, few research has been conducted at video-level, which requires further exploration. 

Inspired by this, we aim at learning more universal representations of temporal inconsistency for Deepfake video detection, which is based on a newly designed unearthing-common-inconsistency (UCI) framework. Our framework employs a 3D convolution network as backbone to extract the consisitency representation in a common space, then utilizes a contrastive learning strategy to capture the discrepancy of temporal consistency between real videos and fake ones from multiple domains. In addition, as aforementioned, CNN detectors are prone to overfitting a domain-specific bias during training, so we assume that a 3D convolution network may inevitably learn spatial domain bias during the convolution process along spatial channels. To this end, we design a task-specific data augmentation, preventing our model from learning spatial texture and preserving the temporal information along temporal channel. As demonstrated in the experiments, this method effectively improves the generalisability across domains. We test our proposed method on public datasets and it surpasses video analysis baselines and state-of-the-art Deepfake detectors, in terms of detection performance across different datasets, confirming the validity of our method. In conclusion, our main contributions are three-folds:
\begin{itemize}
\item We propose a novel Deepfake video detection by unearthing temporal  inconsistency clue that commonly exists in different manipulation techniques. A contrast learning strategy is adopted for better domain generation.
\item We extract the temporal representation in a common space for both real and fake videos through a weight shared network and focus our model on temporal information by applying a task-specific temporally-preserved augmentation. Ablation studies prove the effectiveness of such design.
\item We conduct comprehensive evaluations on several benchmarks and demonstrate the superior generalisability of the proposed model.
\end{itemize}
\vspace{0.5cm}
\section{Related Work}

\subsection{Deepfake Detection}
Recent Deepfake detectors mainly attempt to mine space-aware or frequency-aware clues in fake videos. Dang et al. \cite{stehouwer2019detection}leverage an attention mechanism attention maps to highlight the informative regions for improving the detection ability. Wang et al. \cite{wang2022fake} use semantic masks as an attention-based data augmentation module to guide detectors focus on forged region. Binh and Woo \cite{woo2022add} explore applications of frequency domain learning and optimal transport theory in knowledge distillation to improve the detection performance of low-quality compressed Deepfakes images. Interestingly, some works \cite{huang2023implicit,dong2023implicit} consider identity information as auxiliary to facilitate binary classifiers. Besides, a series of approaches \cite{qian2020thinking,wang2023dynamic,miao2022hierarchical}analyse images in frequency domain, a vital method wildly used in image classification and steganalysis \cite{chen2017jpeg,denemark2016steganalysis}, thereby improving detection robustness.

\subsection{Generalisable Method}
Although a relative high accuracy can be achieved when detectors are trained and tested on a similar distribution, it is still a challenge to overcome performance decline on unseen forgeries with distinct domain bias. To solve this issue, Li et al. \cite{li2020face} uses a self-supervised learning strategy to predict the blending boundaries caused by the common post-processing shared by forgery procedures. Basing on the meta-learning strategy, Sun et.al \cite{sun2021domain} assign different sample with adaptive weights to balance the model’s generalization across multiple domains. Dong et.al \cite{dong2023implicit} propose the Multi-scale Detection Module that diminishes the unexpected learned identity representation on images, which is proven to be an obstacle for generalization.

Another practical approach is to excavate the short-term or long-term temporal inconsistency in fake videos. Since the majority of manipulations render target faces in a frame-by-frame manner, without introducing temporal contexts, this may inevitably ruins the consistency of original videos and leaves subtle clues for detectors. For instance, Haliassos et al. \cite{haliassos2021lips} finetune a temporal network pre-trained on lipreading task to learn high-level semantic irregularities in mouth movements. Zhao et al. find a strong correlation between audio and lip movement in speech videos. They extract generic representations of audio-visual information, then use a self-supervised pre-trained framework to achieve better accuracy and generalization. In light of local motion in snippets,  Gu et al. design an Intra-Snippet Inconsistency Module and an Inter-Snippet Interaction Module as complementary components to detect dynamic inconsistency in Deepfake videos. In summary, these methods model the inconsistency that unfeasible to be fixed by generative models at this stage, making it a possible way to explore more generalisable detectors.

\subsection{Contrastive Learning}
Contrastive learning has gained significant attention in recent years due to its success in downstream tasks, such as classification, clustering, and retrieval \cite{qian2021spatiotemporal}. The central idea behind contrastive learning is to pull together similar data samples while pushing apart dissimilar ones in a high-dimensional space. This encourages the model to capture inherent features or representations that can effectively discriminate between different samples. Recently, many approaches deploy a contrastive learning strategy to help the model capture more discriminative feature, resulting in better generalization of the models. Examples like Fung et al. \cite{fung2021deepfakeucl,dong2023contrastive} and Sun et al. \cite{sun2022dual}, they integrate contrastive learning with Deepfake detection task, and design task-specific sample pairs using data augmentation methods, boosting the unseen domain performance.

Inspired by the above works, we also use contrastive learning to extract temporal inconsistency representations in a supervised manner. Accordingly, a temporal-preserved augmentation is carefully devised, and we argue that this could refrain the model from learning redundant information except for temporal representations.

\section{Proposed Method}
\subsection{Overall Framework}
We first introduce our proposed Unearthing Common Inconsistency (UCI) framework for Deepfake video detection, which could induce the general temporal inconsistency in forgery videos from different domains. Specifically. we extract the representation of real videos and fake videos via a weight-shared temporal network and train the model in a supervised contrastive learning manner. Additionally, a temporal-preserved augmentation Module is carefully designed to augment these video clips only in the RGB plain. This could further facilitate the extraction of high-dimensional temporal representations. Eventually, these distinct representations undergo an attention-based Consistency Correlation Learning Module to fully analyse the variance between sample pairs with different labels. The framework of our method is shown in Figure~1.

\subsection{Video Encoder}
We extract the temporal representation using Inflated 3D ConvNet (I3D) \cite{}. I3D inflates all the filters and pooling kernels from a 2D ConvNet architecture, demonstrating robust performance and transferability in multiple action recognition tasks. Each video clip is mapped into a 2048-dimensional representation to extract the underlying long-term sequential dependency. Since our method is plug-and-play and can integrate into existing models, we also replace I3d with other video analyse networks as encoder backbone to test the effectiveness and versatility of our approach.

\begin{algorithm}[tb]
\caption{Temporal-preserved augmentation}
\label{alg:algorithm}
\textbf{Input}: Video clip $X = \{f_1,f_2,\cdots,f_N \}$ with N frames\\
\textbf{Resize}: Resize to size of 224 × 224\\
\textbf{Crop}: Randomly crop a spatial region for all the frames with same size ratio \textbf{S} in range of [0.8, 1] and same aspect ratio \textbf{A} in [0.75, 1.3]. Draw a flag $\textbf{P}_c$ with $20 \%$ on 1\\
\textbf{Blur}: Randomly Gaussian blur all the frames. Draw a flag $\textbf{P}_b$ with $10 \%$ on 1\\
\textbf{Flip}: Randomly flip all the frames. Draw a flag $\textbf{P}_f$ with $50 \%$ on 1\\
\textbf{Vertical flip}: Randomly vertically flip all the frames. Draw a flag $\textbf{P}_v$ with $50 \%$ on 1\\
\textbf{Color jitter}: Randomly color jitter. Draw a flag $\textbf{P}_{cj}$ with $70 \%$ on 1\\
\textbf{Greyscale}: Randomly greyscale. Draw a flag $\textbf{P}_g$ with $70 \%$ on 1\\
\textbf{Cutout}: Randomly cutout a square region with side length \textbf{L} in range of [32, 64]. Draw a flag $\textbf{P}_{co}$ with $70 \%$ on 1

\begin{algorithmic}[1] 
\STATE $X$=Crop($X$, $size$=\textbf{S}, $aspect$=\textbf{A}) if $\textbf{P}_c = 1$
\STATE $X$=Blur($X$) if $\textbf{P}_b = 1$
\STATE $X$=Flip($X$) if $\textbf{P}_f = 1$
\STATE $X$=Vertical\_flip($X$) if $\textbf{P}_v = 1$
\FOR{$i$ in \{1,\dots,N\}}
\STATE $f^{\prime}_i$ = Resize($f_i$)
\STATE $f^{\prime}_i$ = Color\_jitter($f^{\prime}_i$) if $\textbf{P}_{cj} = 1$
\STATE $f^{\prime}_i$ = Greyscale($f^{\prime}_i$) if $\textbf{P}_g = 1$
\STATE $f^{\prime}_i$ = Cutout($f^{\prime}_i$, $length$=\textbf{L}) if $\textbf{P}_{co} = 1$
\ENDFOR \\
\end{algorithmic}
\textbf{Output}: Augmented video clip $X^{\prime}$=$\{f^{\prime}_1,f^{\prime}_2,\cdots,f^{\prime}_N \}$
\end{algorithm}

\subsection{Temporal-Preserved Augmentation}
It is nature to generate different views of samples in a contrastive learning method \cite{he2020momentum}. This could not only direct the model's attention towards more salient features, but also pull closer samples in the same label while with different bias, resulting in better generalization. Previous works utilize some common augmentation techniques, such as random clipping, horizontal flipping, and Gaussian noise in image level \cite{wang2021representative,chen2021local,sun2022dual}, as well as frame shuffle and playback rates altering in video level \cite{lee2017unsupervised}. However, directly incorporating these augmentations into our task would ruin the temporal consistency of original videos. Unlike related works \cite{de2020deepfake,sun2022dual}, we divide augmentation techniques into two groups. The first type only introduces local spatial randomness and does not break the motion cues across frames, which could be applied on each frame with independent probability, such as random cutoutting, greyscaling and color jittering. On the contrary, the other one includes random flip, vertical flip, cropping and blurring, which needs to be performed on all the frames to maintain temporal coherence. Table~5 illustrates the effectiveness of this approach. Algorithm~\ref{alg:algorithm} elaborates the detailed process of the temporal-preserved augmentation.

\begin{figure*}[t]
\centerline{\includegraphics[width=7in]{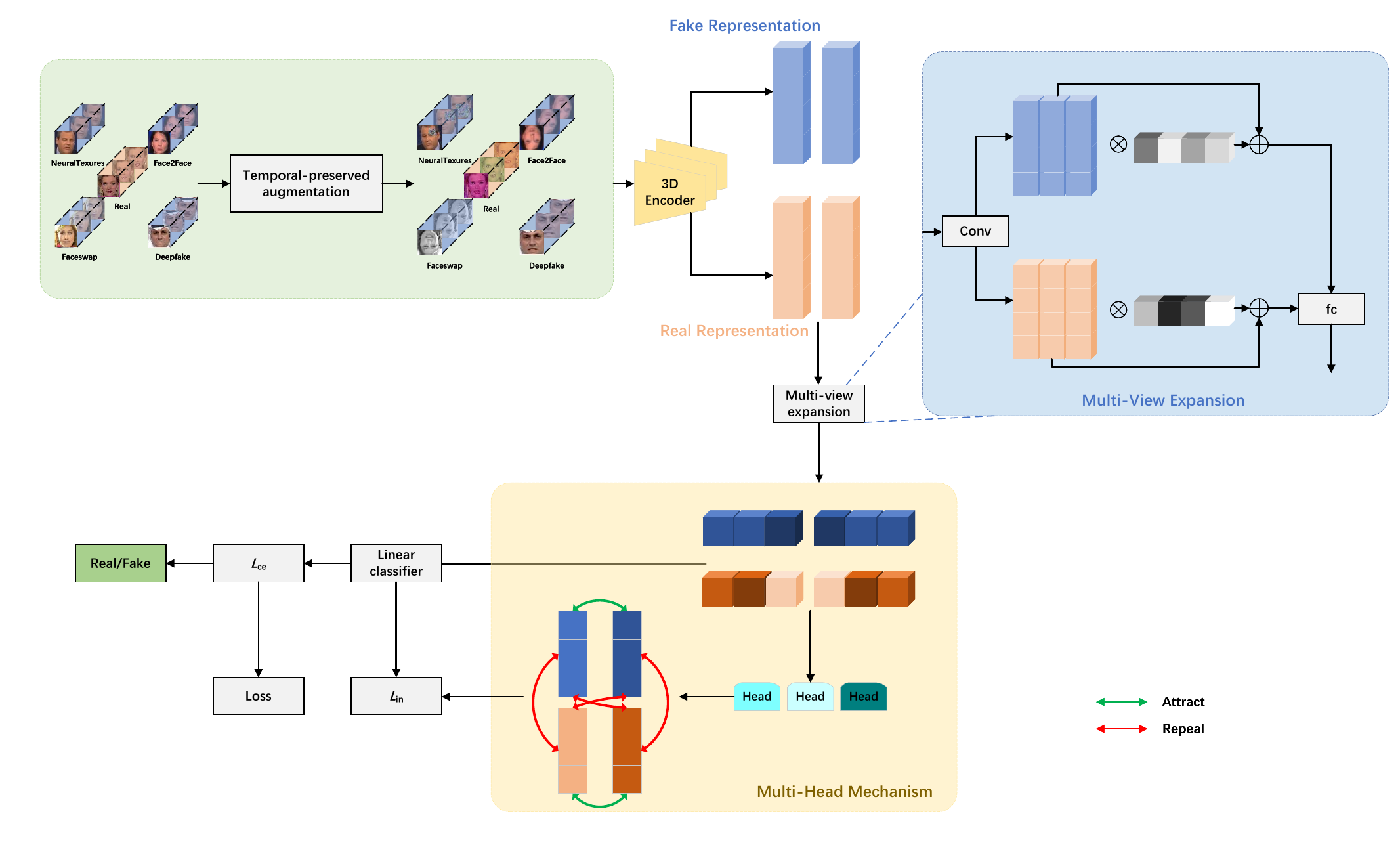}}
\caption{Illustration of our proposed Unearthing Common Inconsistency (UCI). First, we input genuine videos along with forged samples from multiple domains. Through a temporal-preserved augmentation, we maintain temporal consistency while disrupting spatial information to encourage the model's emphasis on temporal features. Subsequently, a temporal convolutional encoder is employed to extract high-dimensional video representations. This is followed by a multi-view expansion module, which captures temporal features of the representations from various perspectives. Finally, a multi-head attention mechanism, combined with a contrastive learning strategy, is applied. This serves to reduce the distance between representations of the same class while increasing the distance between negative pairs of samples, facilitating enhanced differentiation of fake videos. $\oplus$ denotes element-wise addition and $\otimes$ denotes channel-wise multiplication.}
\end{figure*}

\vspace{0.2cm}
\subsection{Attention-Based Interaction}
As already mentioned, a temporal network acts as the encoder to extract the high-dimensional representation of consistency for each video. Hence, the essence of enabling our approach to discern between genuine and fake videos lies in how to effectively differentiate subtle distinctions among representations. If we employ the prior methods \cite{zhao2022self,qian2021spatiotemporal}  by utilizing directly high-dimensional representations as inputs to the loss function, it could lead to significant temporal information loss, severely compromising detection performance. To address this challenge, we devise a novel Attention-Based Consistency Correlation Learning module specifically for temporal representations, introducing diverse information through different views. Additionally, an interaction module based on the multi-head attention mechanism is integrated, enabling the discernment of both similarities and differences in long-range dependencies among representations of genuine and fake samples. 
\vspace{0.1cm}
\subsubsection{Multi-View Expansion}
In order to extract locally and globally rich intrinsic features for each representation, we initiate the process by expanding the multi-view content of representations through a convolutional layer. Inspiration by SENet \cite{hu2018squeeze}, we enhance feature representation by learning view-wise relationships and adaptively recalibrating feature maps, as shown in Figure~2. This mechanism enables the network to allocate more importance to informative views while suppressing less relevant ones, resulting in improved discriminative power and enhanced generalization across domains. 

Formally, let $I \in R^{2048 \times 1}$ denotes an encoded representations of an augmented video clip. First, we expand temporal views using a convolutional layer and obtain a multi-view representation $I_{mv} \in R^{2048 \times 512}$. Then, a compress-and-restore operation along the original representation direction are applied by two fully-connection layers $fc_c$ and $fc_r$ respectively, with a compression ratio $r$, obtaining the weight of different temporal views:
\begin{equation}
    W_{se} = Sigmoid(fc_r(fc_c(GAP(I_{mv}^{T}), r)))\label{e1},
\end{equation}
where $GAP$ represents global average pooling and $Sigmoid$ refers to sigmoid function. Then we perform channel-wise multiplication on $W_{se}$ and $I_{mv}^{T}$, resulting in a weighted temporal representation map. Subsequently, a residual connection \cite{he2016deep} is introduced to prevent information loss and gradient vanishing. Finally, through a fully connected layer $fc$, a comprehensive representation containing enriched multi-view information is obtained:
\begin{equation}
    Z = fc(I_{mv}^{T} \oplus (I_{mv}^{T} \otimes W_{se} )),
\end{equation}
where $\oplus$ denotes element-wise addition and $\otimes$ denotes channel-wise multiplication.

As a common practice in Deepfake detection task, the classifier is required to output binary values to make the final determination of the label for input video. Following this, to conduct classification and leverage label information effectively, a fully connected classifier $f_{final}$ is added subsequent to the enrichment of representations. The binary cross-entropy loss $\mathcal{L}_{ce}$ is expressed as:
\begin{equation}
    \mathcal{L}_{ce} = y\log y^{\prime} + (1-y)\log(1-y^{\prime}),
\end{equation}
where $y$ denotes the authentic label, $y^{\prime}$ is the final predicted probability. 
\vspace{0.5cm}
\subsubsection{Multi-Head Mechanism}
We devise a task-oriented multi-head attention mechanism, aimed at effectively integrating diverse dependency relationships among representations. Given a representation $Z \in R^{512 \times 1}$, we assign $n$ heads with $n$ learnable convolutional projection weights $\{w_i|i \in (1,n)\}$. Then the attention interaction between representations can be calculated by
\begin{equation}
    head_i=Softmax(\frac{w_i(Z)(w_i(Z^{\prime}))^{T}}{\sqrt{d}}),
\end{equation}
and
\begin{equation}
    Att(Z, Z^{\prime}) = Concat(head_1, ..., head_n)\label{e2},
\end{equation}
where $Z^{\prime}$ represents another video representation, $\sqrt{d}$ acts a normalization factor to avoid value explosion, $d$ denotes the dimension size of the representation.

\subsection{Loss Function}
Adhering to the principles of contrastive learning, we employ the InfoNCE \cite{oord2018representation} loss on the processed representations. Give a representation set of real clips $Z_r \in \{z_{r1},z_{r2},\dots,z_{rn}\}$ and a representation set of fake clips $Z_f \in \{z_{f1},z_{f2},\dots,z_{fn}\}$, the loss is calculated based upon INfoNCE as follows:
\begin{equation}
    \mathcal{L}_r=-\log \frac{\sum_{i \neq j}{e^{Att(z_{ri},z_{rj})/\tau}}}{\sum_{i \neq j}{e^{Att(z_{ri},z_{rj})/\tau}}+\sum_i{\sum_j{e^{Att(z_{ri},z_{fj})/\tau}}}},
\end{equation}
\begin{equation}
    \mathcal{L}_f=-\log \frac{\sum_{i \neq j}{e^{Att(z_{fi},z_{fj})/\tau}}}{\sum_{i \neq j}{e^{Att(z_{fi},z_{fj})/\tau}}+\sum_i{\sum_j{e^{Att(z_{fi},z_{rj})/\tau}}}},
\end{equation}
and
\begin{equation}
    \mathcal{L}_{in}=\frac{1}{2}\mathcal{L}_r+\frac{1}{2}\mathcal{L}_f,
\end{equation}
where $\tau$ is the temperature which is set 0.1. The overall loss function is formulated as:
\begin{equation}
    \mathcal{L}=\alpha \mathcal{L}_{in}+(1-\alpha)\mathcal{L}_{ce}\label{e3},
\end{equation}
where $\alpha$ is the hyper-parameter used to balance the contrastive loss and cross-entropy loss.

\begin{figure}[t]
\centerline{\includegraphics[width=3in]{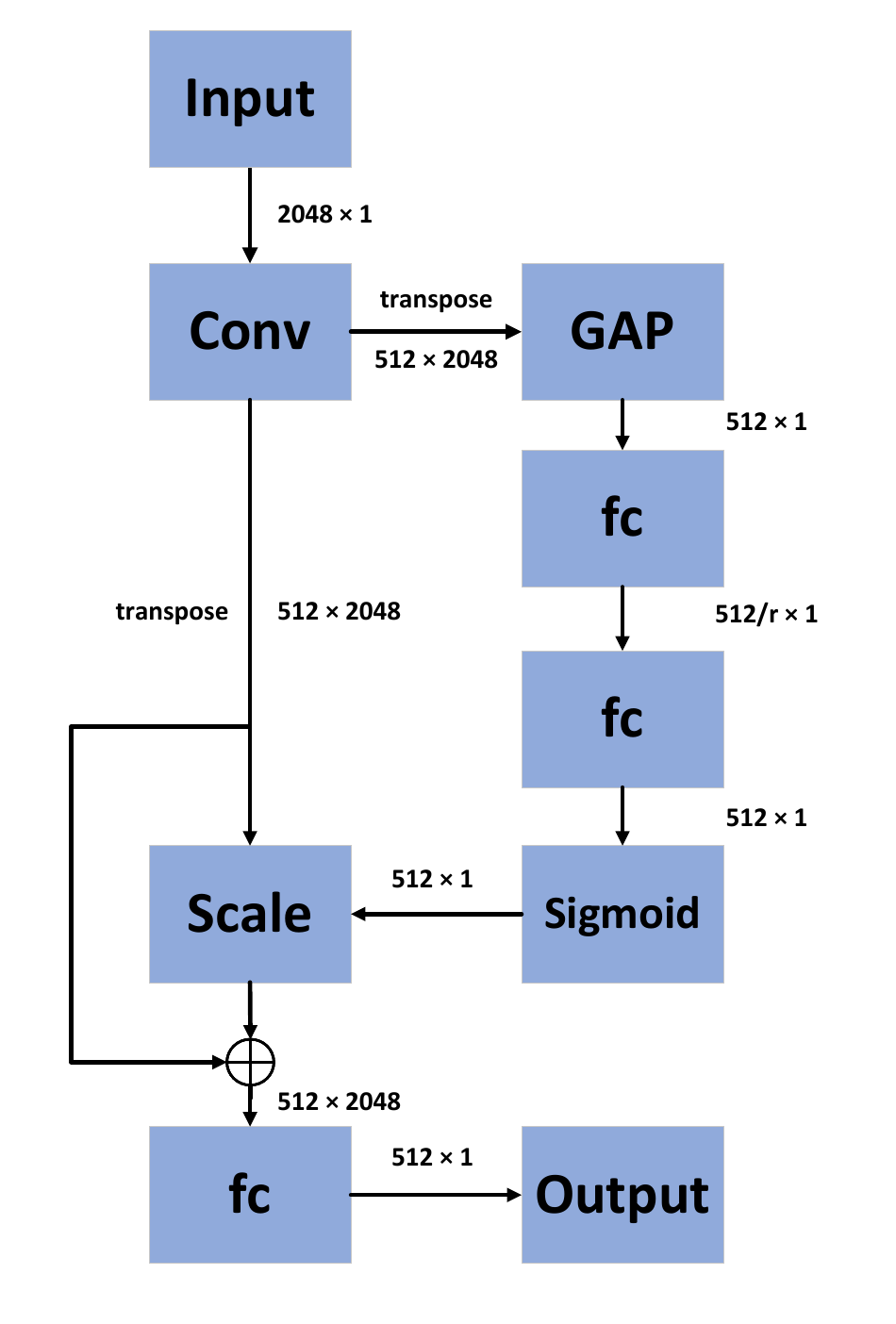}}
\caption{Illustration of the Multi-View Expansion module.}
\end{figure}
\vspace{0.5cm}

\vspace{0.5cm}
\begin{table*}[]
\begin{center}
\begin{tabular}{c|cccccccccc}
\hline
\multirow{2}{*}{Method} &
  \multicolumn{10}{c}{Training on remaining three} \\ \cline{2-11} 
 &
  \multicolumn{2}{c|}{Deepfake} &
  \multicolumn{2}{c|}{FaceSwap} &
  \multicolumn{2}{c|}{Face2Face} &
  \multicolumn{2}{c|}{NeuralTexture} &
  \multicolumn{2}{c}{Avg} \\ \hline
 &
  AUC &
  \multicolumn{1}{c|}{ACC} &
  AUC &
  \multicolumn{1}{c|}{ACC} &
  AUC &
  \multicolumn{1}{c|}{ACC} &
  AUC &
  \multicolumn{1}{c|}{ACC} &
  AUC &
  ACC \\
LSTM &
  88.4 &
  \multicolumn{1}{c|}{77.5} &
  85.3 &
  \multicolumn{1}{c|}{75.2} &
  85.9 &
  \multicolumn{1}{c|}{76.2} &
  84.7 &
  \multicolumn{1}{c|}{74.5} &
  86.1 &
  75.8 \\
C3D &
  88.3 &
  \multicolumn{1}{c|}{77.0} &
  84.0 &
  \multicolumn{1}{c|}{72.4} &
  81.3 &
  \multicolumn{1}{c|}{72.3} &
  83.7 &
  \multicolumn{1}{c|}{72.2} &
  84.3 &
  73.5 \\
MS-TCN &
  83.0 &
  \multicolumn{1}{c|}{71.0} &
  83.4 &
  \multicolumn{1}{c|}{73.3} &
  88.2 &
  \multicolumn{1}{c|}{77.3} &
  85.8 &
  \multicolumn{1}{c|}{74.3} &
  85.1 &
  74.0 \\ \hline
EfficientNet &
  82.9 &
  \multicolumn{1}{c|}{72.6} &
  81.3 &
  \multicolumn{1}{c|}{69.6} &
  84.3 &
  \multicolumn{1}{c|}{74.7} &
  79.6 &
  \multicolumn{1}{c|}{78.1} &
  82.0 &
  73.7 \\
Mesconet &
  89.2 &
  \multicolumn{1}{c|}{79.2} &
  85.4 &
  \multicolumn{1}{c|}{75.1} &
  83.0 &
  \multicolumn{1}{c|}{71.7} &
  82.4 &
  \multicolumn{1}{c|}{74.1} &
  85.0 &
  75.0 \\
Lipforensics &
  92.3 &
  \multicolumn{1}{c|}{\textbf{83.8}} &
  87.3 &
  \multicolumn{1}{c|}{77.9} &
  93.0 &
  \multicolumn{1}{c|}{82.9} &
  84.4 &
  \multicolumn{1}{c|}{72.6} &
  89.2 &
  79.3 \\
F3-net &
  \textbf{92.4} &
  \multicolumn{1}{c|}{82.4} &
  \textbf{90.5} &
  \multicolumn{1}{c|}{\textbf{81.8}} &
  92.2 &
  \multicolumn{1}{c|}{81.1} &
  86.5 &
  \multicolumn{1}{c|}{75.8} &
  90.4 &
  80.3 \\
Capsule &
  87.8 &
  \multicolumn{1}{c|}{76.2} &
  83.9 &
  \multicolumn{1}{c|}{75.3} &
  86.1 &
  \multicolumn{1}{c|}{76.5} &
  86.7 &
  \multicolumn{1}{c|}{77.4} &
  86.1 &
  76.3 \\
multi-task &
  86.9 &
  \multicolumn{1}{c|}{76.8} &
  82.5 &
  \multicolumn{1}{c|}{70.9} &
  85.0 &
  \multicolumn{1}{c|}{76.4} &
  86.3 &
  \multicolumn{1}{c|}{76.7} &
  85.2 &
  75.2 \\
RECCE &
  84.7 &
  \multicolumn{1}{c|}{75.9} &
  86.4 &
  \multicolumn{1}{c|}{74.7} &
  88.3 &
  \multicolumn{1}{c|}{78.2} &
  82.4 &
  \multicolumn{1}{c|}{73.6} &
  85.4 &
  75.6 \\ \hline
UCI (ours) &
  92.3 &
  \multicolumn{1}{c|}{81.5} &
  88.9 &
  \multicolumn{1}{c|}{78.5} &
  \textbf{93.2} &
  \multicolumn{1}{c|}{\textbf{83.9}} &
  \textbf{87.9} &
  \multicolumn{1}{c|}{\textbf{78.6}} &
  \textbf{90.6} &
  \textbf{80.6} \\ \hline
\end{tabular}
\label{tab1}
\caption{Video-level generalization tests accuracy (\%) and AUC scores (\%) within FF++.}
\end{center}
\end{table*}

\section{Experiments}
\subsection{Experimental Settings}
\subsubsection{Datasets}
We evaluate our method on the widely-used benchmark dataset FaceForensics++ \cite{rossler2019faceforensics++}. FF++ contains 1000 original videos and 4000 a fake videos forged by four manipulation methods, i.e. Deepfakes \cite{DeepF}, Face2Face \cite{thies2016face2face}, FaceSwap \cite{DeepFakes} and NeuralTextures \cite{thies2019deferred}, yielding 5000 videos in total. Besides, it also provide multiple video quality, i.e. raw quality (raw) without visual loss, high quality (c23) with minor visual loss and low quality (c30) with heavy visual loss. Furthermore, we also test out method on other three popular datasets, i.e. Celeb-DF \cite{li2020celeb}, DFDC-preview \cite{dolhansky2019deepfake} and FaceShifter \cite{DBLP:journals/corr/abs-1912-13457}, to evaluate the generalization of our approach.

\subsubsection{Baseline Methods}
To validate the effectiveness and transferability of our approach, we compare it with several representative works in Deepfake detection and video analysis. For face forgery detection, we choose EfficientNet \cite{tan2019efficientnet}, Mesconet \cite{Afchar_2018}, Lipforensics \cite{haliassos2021lips}, F3-net \cite{qian2020thinking}, Capsule \cite{nguyen2019use}, multi-task \cite{nguyen2019multitask} and RECCE \cite{cao2022end}. For video analysis, LSTM \cite{graves2012long}, C3D \cite{tran2015learning} and MS-TCN \cite{martinez2020lipreading} are chosen. To ensure equitable comparison, we adhere to the approach outlined in \cite{haliassos2021lips}, whereby we calculate video-level metrics for all models. This involves averaging the model's predictions—each prediction corresponds to either a frame or a video clip—across the entirety of the video for a comprehensive assessment. The state-of-the-art baseline models with source codes published for comparative tests are trained and tested on the same datasets as ours while maintaining their original optimal experiment settings when applicable.

\subsubsection{Implementation Details}
We use Retinaface \cite{deng2020retinaface} to detect and crop faces for all the datasets, then resize them to 224 × 224. Each video clip contains 96 frames. The Kinetics-400 \cite{kay2017kinetics} pre-trained I3D \cite{carreira2017quo} is used as our backbone and the weights of attention heads are randomly initialized. We use a batch size of 16 and Adam \cite{kingma2014adam} optimisation with a learning rate of $10^{-4}$. The head number in Equation~(\ref{e2}) is set to 8, with head dimension 64. The compression rate $r$ in Equation~(\ref{e1}) is set to 4 and balance factor $\alpha$ in Equation~(\ref{e3}) is set to 0.1 for the first 5 epochs as warm-up aiming at binary classification, then set to 0.5.

\begin{table}[]
\begin{center}
\begin{tabular}{cccc|c}
\hline
\multirow{2}{*}{Method} & \multirow{2}{*}{Celeb} & \multirow{2}{*}{DFDC-pre} & \multirow{2}{*}{FShr} & \multirow{2}{*}{Avg} \\
             &      &      &      &      \\ \hline
LSTM         & 67.3 & 58.9 & 81.2 & 69.1 \\
C3D          & 64.2 & 53.5 & 83.6 & 67.1 \\
MS-TCN        & 72.6 & 57.7 & 79.9 & 70.1 \\ \hline
EfficientNet & 59.8 & 47.8 & 82.3 & 63.3 \\
Mesconet     & 62.3 & 56.7 & 86.9 & 68.6 \\
Lipforensics & 74.2 & 68.5 & 93.4 & 78.7 \\
F3-net       & 67.2 & 61.4 & 91.6 & 73.4 \\
Capsule      & 64.5 & 65.8 & 87.5 & 72.6 \\
multi-task   & 75.7 & 68.1 & 86.7 & 76.8 \\
RECCE        & 73.5 & 62.0 & 83.5 & 73.0 \\ \hline
UCI (ours)              & \textbf{77.9}          & \textbf{70.3}                 & \textbf{93.6}                & \textbf{80.6}        \\ \hline
\end{tabular}
\label{tab2}
\caption{Video-level generalization tests AUC scores (\%) on the testing datasets after trained on FF++.}
\end{center}
\end{table}

\vspace{0.5cm}
\subsection{Cross-domain Evaluation within FF++}
In this section, we conduct our experiments on four sub-datasets within FF++. First, we train the proposed UCI model with training set of three datasets and then assess the generalization ability by testing the model on the testing set of the remaining set.

According to Table~1, for Deepfake and FaceSwap, our method ranks among the top three out of other methods, and is comparable to the SOTA F3-net and Lipforensics with decent drop, which may because the uniqueness of the dataset results in inconspicuous inter-frame inconsistencies. For Face2Face and NeuralTextures, our method achieves the best performance in terms of both AUC and ACC. On average, our method outperforms the others in two metrics as well. This indicates that our model possesses strong generalization ability, since temporal inconsistencies are widely present in manipulated videos, and our method could effectively captures them, indicating its generalization capability.
\vspace{0.2cm}
\subsection{Cross-domain Evaluation across Datasets}
In this section, we conduct our experiments on three datasets (Celeb, DFDC-preview and FaceShifter), to further evaluate the generalization ability in a more open scenario, which aligns better with real-world situation. We trained the model using FF++ and test on other datasets.

As illustrated in Table~2, DFDC-preview is observed to be the hardest dataset because it is crafted by 8 different facial manipulation techniques with much more complex scenarios. From Table~2 we can see that, the highest AUC score is all achieved by the our UCI method with a score of $77.9\%$, $70.3\%$ and $93.6\%$, respectively, followed respectively by multi-task with a score of $75.7\%$, $68.1\%$ and $83.5\%$. Taking the average of the AUC scores across the three datasets, our UCI method attains the highest average AUC score of $80.6\%$, which is the only one to achieve the highest AUC score over $80\%$ against all other comparative baseline methods. This signifies that the model has extrapolated a universal temporal consistency representation from domains within FF++, which can generalize equally well across other datasets, leading to better generalization performance.

\begin{table*}[t]
\begin{center}
\begin{tabular}{c|cc|cc}
\hline
\multirow{2}{*}{}                   & \multicolumn{2}{c|}{Celeb}    & \multicolumn{2}{c}{DFDC-pre} \\ \cline{2-5} 
                                      & AUC  & ACC  & AUC  & ACC  \\ \hline
w/o augmentation                      & 74.6 & 63.1 & 66.6 & 58.9 \\
Augmentation w/o temporal-persistence & 69.1 & 60.4 & 62.4 & 54.3 \\
Augmentation w temporal-persistence & \textbf{77.9} & \textbf{69.4} & \textbf{70.3}   & \textbf{68.7}  \\ \hline
\end{tabular}
\label{tab5}
\caption{Ablation study on settings of augmentation module.}
\end{center}
\end{table*}

\begin{table}[]
\begin{center}
\begin{tabular}{l|ll}
\hline
backbone          & AUC           & ACC           \\ \hline
LSTM              & 67.3          & 54.0            \\
LSTM+Ours         & 76.2          & 66.3          \\ \hline
C3D               & 64.2          & 52.6          \\
C3D+Ours          & 72.5          & 61.7          \\ \hline
I3D               & 67.5          & 46.9          \\ 
\textbf{I3D+Ours} & \textbf{77.9} & \textbf{64.7} \\ \hline
\end{tabular}
\label{tab3}
\caption{Ablation study on UCI with video analyse backbones.}
\end{center}
\end{table}

\vspace{0.5cm}
\subsection{Ablation Study}
In this section, we meticulously investigate diverse combinations and individual constituents of the proposed UCI through a series of ablation studies. It's worth emphasizing that all ensuing experiments are trained on the FF++ dataset to ensure the validity of our findings.
\vspace{0.5cm}

\subsubsection{Study on different backbone.}
In our approach, a temporal convolutional network is utilized as an encoder in our approach, responsible for extracting temporal representations from the samples. To demonstrate that the outstanding performance of our approach is not solely contingent on the choice of encoder, we integrate other two temporal convolutional networks into our framework, namely LSTM and C3D. To substantiate that the performance of the proposed approach is actually achieved by the design of the modules implemented. The test is conducted on the CelebDF. As shown in Table~3, it is evident that when integrate UCI into the backbone networks, models that previously exhibited relatively bad performance have observed an enhancement of approximately $10\%$. This provides evidence of UCI's robust transferability across a spectrum of networks, thereby confirming its exceptional capacity for seamless migration while reinforcing the rational foundation of its design.
\vspace{0.5cm}
\begin{table}[]
\begin{center}
\begin{tabular}{c|c|cc}
\hline
\begin{tabular}[c]{@{}c@{}}Temporal-preserved \\ augmentation\end{tabular} &
  \begin{tabular}[c]{@{}c@{}}Contrastive \\ learning\end{tabular}   & AUC           & ACC           \\ \hline
                                &                          & 70.7          & 57.2          \\ 
{\checkmark}        &                          & 73.2          & 62.8          \\ 
                                & \checkmark & 75.6          & 65.1          \\ 
{\checkmark}        & {\checkmark} & \textbf{77.9} & \textbf{69.4} \\ \hline
\end{tabular}
\label{tab4}
\caption{Ablation study on combinations of components.}
\end{center}
\end{table}

\subsubsection{Study on different setting in components.}
To demonstrate the positive influences of our temporal-preserved augmentation, four settings are constructed and compared in Table~4. According to the results, temporal-preserved augmentation can bring a $2.5\%$ gain to AUC and a $5.6\%$ gain to ACC. When contrastive learning is integrated, the model is further improved by $4.9\%$ in AUC and a $7.9\%$ gain to ACC. Finally, when temporal-preserved augmentation and contrastive learning are equipped together, UCI achieves best performance. This indicates that these two modules can effectively collaborate, contributing collectively to the improvement of generalization performance in Deepfake detection.
\vspace{0.7cm}
\subsubsection{Study on different setting in temporal-preserved augmentation.}
Table~5 presents the performance metrics (AUC and ACC) for two datasets, Celeb and DFDC-preview, under three different experimental conditions:

"w/o augmentation" refers to the case where no augmentation was applied. "Augmentation w/o temporal-persistence" indicates that randomly applies all the augmentation to each frames, including augmentations that can ruin temporal consistency. The results show a decrease in performance compared to the first condition. This may such augmentation break the consistency between frames, resulting in even worse performance. "Augmentation w temporal-persistence" represents that while introducing augmentations that preserve temporal consistency, we also impose constraints on augmentations that disrupt temporal consistency, as aforementioned in the detail of temporal-preserved augmentation. This configuration yielded the highest performance among all the conditions. This can be concluded that incorporating both augmentation and temporal persistence leads to the best overall performance. This demonstrates the rationale behind the design of temporal-preserved augmentation.

\vspace{0.7cm}
\section{Conclusion}
In this research, we delve into the realm of enhancing the generalization capability of Deepfake detection by addressing the common inconsistency prevalent in manipulated videos. Our focus centers on introducing a novel temporal-preserving augmentation methodology, steering the detector towards probing temporal representations as opposed to spatial artifacts. Additionally, an interaction module employing attention-based mechanism and contrastive learning is incorporated to further elevate performance standards. Furthermore, the comprehensive array of experiments underscores the efficacy of this design in effectively capturing the similarities and discrepancies in representations between authentic and fabricated videos. This not only points to superior generalization potential across multiple datasets but also positions itself as a more adept solution when compared to existing methodologies.

\newpage
\bibliography{aaai24}

\end{document}